\documentclass[lettersize,journal]{IEEEtran}
\usepackage{amsmath,amsfonts}
\usepackage{algorithmic}
\usepackage{algorithm}
\usepackage{array}
\usepackage[caption=false,font=normalsize,labelfont=sf,textfont=sf]{subfig}
\usepackage{textcomp}
\usepackage{stfloats}
\usepackage{url}
\usepackage{verbatim}
\usepackage{graphicx}
\usepackage{cite}
\usepackage{xcolor}
\usepackage{multirow}

\begin{document}

\title{Architecture-aware Robustness Evaluation of Explainable Deep Learning for Breast Cancer Diagnosis}

\author{Balenthira Thanusanth, Selvarajah Thuseethan, \IEEEmembership{Member, IEEE},
Roshan G. Ragel, \IEEEmembership{Senior Member, IEEE}, Bimali S. Weerakoon, Ayesh Jayasinghe and Ananthamoorthy Krishnamoorthy
\thanks{This work did not involve human subjects or animals in its research.}
\IEEEcompsocitemizethanks{\IEEEcompsocthanksitem Corresponding Author: S. Thuseethan (thuseethan.selvarajah@cdu.edu.au)
			\IEEEcompsocthanksitem Balenthira Thanusanth, Roshan G. Ragel, Bimali S. Weerakoon and Ayesh Jayasinghe are with the University of Peradeniya, Sri Lanka. Selvarajah Thuseethan is with Charles Darwin University, Australia. Ananthamoorthy Krishnamoorthy is with Aneurin Bevan University Health Board, United Kingdom}
            }

\markboth{Preprint Under Review}%
{Shell \MakeLowercase{\textit{et al.}}: Bare Demo of IEEEtran.cls for IEEE Journals}

\maketitle

\begin{abstract}
Explainable Artificial Intelligence (XAI) has become essential in medical image analysis to ensure transparency of deep learning (DL)-based diagnostic systems. However, selecting appropriate XAI techniques for breast cancer recognition remains largely ad hoc, with limited systematic evaluation across different DL architectures. This study presents a systematic architecture-aware evaluation protocol to assess the effectiveness of nine widely used XAI techniques across four categories of DL models: very deep, lightweight, transformer-based and hybrid neural networks. The evaluation is conducted on a breast ultrasound dataset comprising 780 images using clinically aligned spatial metrics, including Pointing Game, Intersection over Union and Mean Coverage, to quantify agreement between generated explanations and expert-annotated lesion regions. Results indicate that explanation quality is primarily influenced by the interaction between model architecture and XAI method, rather than any single technique consistently outperforming others. Hybrid architectures produce more spatially coherent explanations, while lightweight and transformer-based models exhibit greater variability across methods. The findings show that no single technique generalises across architectures and evaluation criteria, emphasising the need for joint selection of DL models and XAI techniques. Explainability depends on both model design and explanation strategy and should not be considered independently. This work provides a structured evaluation protocol and practical guidance for selecting XAI techniques in breast cancer diagnosis, supporting more transparent clinical decision-support systems. \textcolor{blue}{Code is publicly available at https://github.com/Nishan-Charlie/Explainable-AI}
\end{abstract}

\begin{IEEEkeywords}
Explainable Artificial Intelligence, Deep Learning, Breast Cancer Recognition, Medical Image
\end{IEEEkeywords}

\section{Introduction}
\IEEEPARstart{D}{eep} Learning (DL) methods are increasingly embedded in clinical decision-making owing to their strong predictive performance, robustness and capacity to support diagnostic processes \cite{almadani2025systematic}. DL has demonstrated high efficacy in oncological applications, particularly in breast cancer detection, where early identification is critical for improving survival outcomes \cite{obeagu2024breast}. Despite these advantages, the opaque nature of many DL models raises concerns in clinical settings. In conventional practice, diagnoses are supported by explicit justifications, strengthening patient trust and acceptance. Consequently, Explainable Artificial Intelligence (XAI) has gained importance in addressing the gap between predictive performance and interpretability \cite{alkhanbouli2025role, rajesh2026improving, shafik2026systematic}. In this context, XAI plays a critical role in medical image–based cancer diagnosis, where explainability is essential for reliable clinical adoption.

Recent studies combine advanced DL architectures with XAI methods to highlight image regions relevant to model decisions and assist clinical interpretation \cite{murugan2025efficient, pandey2026advancing, singh2026mammxai}. However, the selection of XAI techniques is often not systematic. Many works rely on commonly used approaches such as SHAP \cite{saharan2025deep, rahman2025enhanced, schindele2025interpretable}, LIME \cite{gupta2025interpretable, ansari2026explainable, abraham2025transparent}, and Grad-CAM \cite{jafarpoor2025enhanced, raju2025medivision, christobel2026novel} without sufficient justification or evaluation of their suitability. As a result, explanations may be technically valid but not aligned with clinical expectations or decision-making needs. Limited involvement of healthcare professionals in validation further reduces practical relevance. In addition, the absence of standardised evaluation frameworks restricts consistent assessment of explanation quality \cite{shifa2025review}. Despite the growing adoption of XAI in breast cancer analysis, comparative evaluation of different techniques remains limited. Existing studies rarely assess clinical utility in a structured manner, and dataset-related challenges continue to be identified as key gaps \cite{jin2022evaluating, fontes2024application, azad2026systematic}. The use of model-agnostic methods such as SHAP is often driven by ease of implementation rather than demonstrated effectiveness. Furthermore, there is no clear guidance on selecting suitable XAI methods for different DL architectures, leading to inconsistent practices and limiting the reliability of decision-support systems. This highlights the need for automated comparative evaluation using standardised metrics to identify effective and trustworthy approaches \cite{kaba2024explainable, ghasemi2024explainable, ansari2025role}.

\subsection{Clinical Deployment Scenario and Limitations}
\begin{figure}[!htbp]
    \centering
    \includegraphics[width=\linewidth]{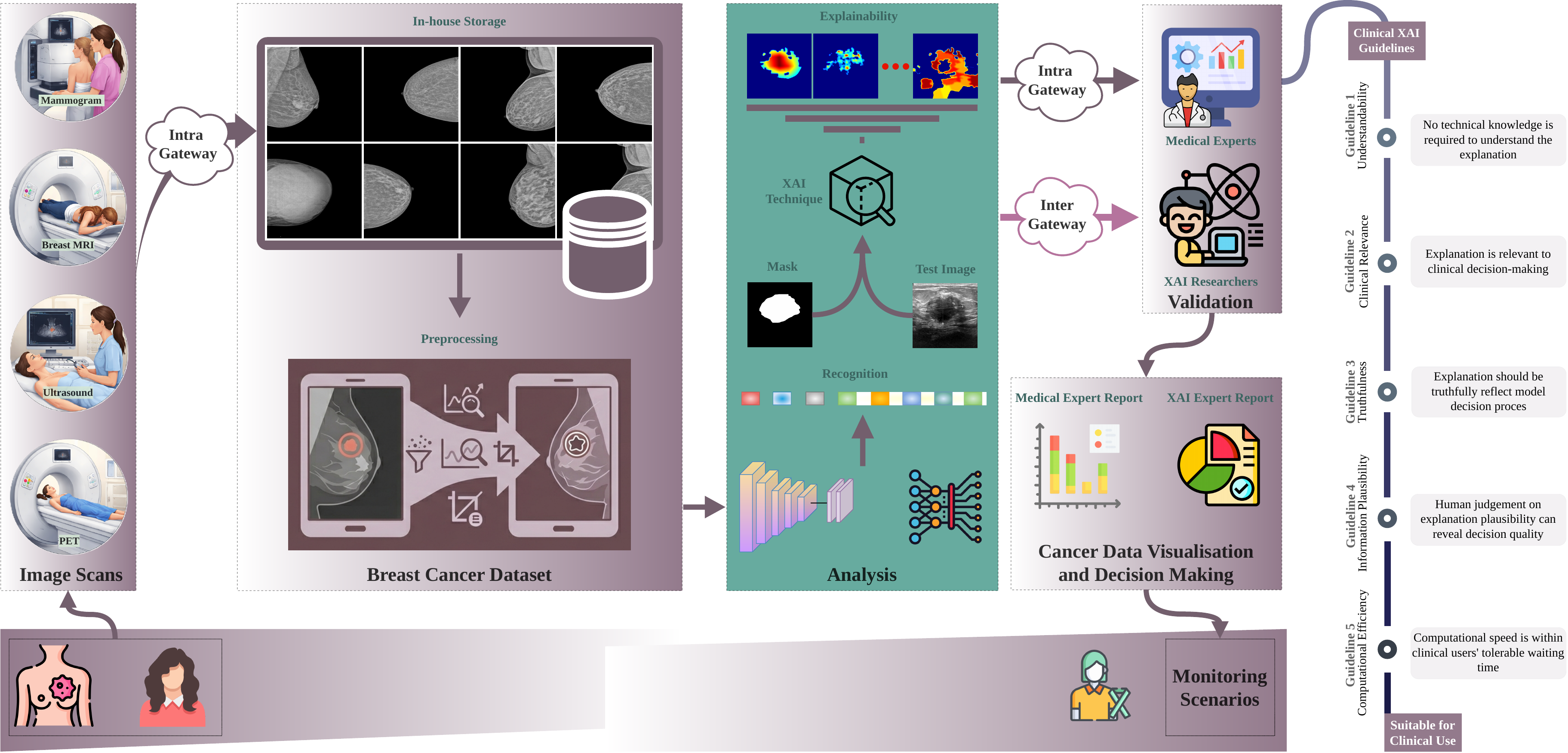}
    \caption{Overview of an end-to-end explainable breast cancer recognition pipeline with clinical utility. The focus of this study is highlighted in the green shaded area.}
    \label{fig:endtoend}
\end{figure}

Consider a comprehensive framework for explainable breast cancer recognition in a clinical setting, as illustrated in Figure \ref{fig:endtoend}. The area highlighted in green represents the focus of this study. XAI outputs are validated locally (i.e., intra-gateway) by medical experts for clinical usefulness, while their technical validity is assessed remotely (i.e., inter-gateway) by XAI researchers. Effective clinical adoption requires satisfying five key criteria outlined in the Clinical XAI Guidelines, namely that explanations are understandable, clinically relevant, truthful, informative, and computationally efficient \cite{jin2023guidelines}. Although ideally all criteria should be met, this is rarely achievable in practice. In automated evaluation settings without expert involvement, explanation quality is instead assessed using quantitative metrics. In summary, the following limitations are identified.

\begin{enumerate}
    \item \textbf{Lack of systematic evaluation:} XAI methods are often applied without rigorous justification or standardised evaluation. As a result, explanations may be technically correct but clinically misaligned.

    \item \textbf{Lack of architecture-aware evaluation:} Selection of XAI methods is typically independent of DL architecture, ignoring their interaction. This leads to inconsistent practices and limits the reliability of XAI-based diagnostics.
\end{enumerate}

To address the limitations of the suboptimal use of XAI techniques with DL in real-world clinical settings, this paper proposes a novel XAI evaluation architecture with automated comparative evaluation using standardised metrics for breast cancer diagnosis. In this study, nine representative and widely used XAI techniques for image classification were taken into consideration. These methods were selected from five broad categories, such as activation-based methods, gradient-based attribution methods, perturbation-based methods, local approximation methods and game-theoretic attribution methods. In the domain of DL, four primary architectural directions have been widely studied, namely very deep neural networks, lightweight neural networks, transformer-based neural networks and hybrid neural networks. Very deep neural networks demonstrate strong representational capacity and have been widely applied to breast cancer detection \cite{gopal2025automatic, nagalakshmi2026leveraging, attallah2026merge}.
In contrast, lightweight neural networks are designed to reduce computational complexity while maintaining competitive performance, which makes them suitable for breast cancer recognition in resource-constrained clinical environments \cite{liu2025rsdcnet, jaafari2026lightweight, kaur2025enhancing}. More recently, transformer-based architectures have attracted considerable attention due to their capacity to model long-range dependencies and global context, which is advantageous in breast cancer diagnosis \cite{naas2025explainable, patheda2025robust, maurya2025bmea}. Hybrid neural networks integrate multiple architectural concepts and demonstrate strong effectiveness in breast cancer recognition \cite{puvar2026deeprnn, purushotham2026breast, sait2026novel}. The key contributions of this work are threefold and are summarised as follows.

\begin{enumerate}
    \item A systematic evaluation of nine representative and widely used XAI techniques for breast cancer recognition is conducted across nine DL models across four architecture categories to systematically assess their suitability for this task. To the best available knowledge, this is among the first studies to provide a comprehensive comparison of benchmark XAI techniques in breast cancer diagnosis.

    \item The performance of the XAI techniques is evaluated using a breast cancer dataset comprising 780 breast ultrasound images across three classes. The evaluation is carried out using three quantitative metrics to assess their effectiveness in this context.

    \item Through comprehensive experiments, an empirical guidance for selecting the most suitable XAI technique for a given type of DL architecture in the context of breast cancer recognition. This facilitates the development of a more effective breast cancer decision-making system in clinical settings.
\end{enumerate}

The rest of this paper is structured as follows. In Section \ref{sec:relatedwork}, the recent literature on explainable breast cancer recognition using DL is reviewed. The proposed method and its contributions are described in Section \ref{sec:proposedmethod}. The experimental results are discussed in Section \ref{sec:experiments}, followed by the conclusion and future works in Section \ref{sec:conclusion}. 

\section{Related Work}
\label{sec:relatedwork}
This section reviews recent advances in breast cancer recognition, focusing on (A) benchmark XAI techniques for image-based analysis and (B) deep learning-based explainable breast cancer recognition.

\subsection{XAI Benchmarks}
XAI techniques for image classification have advanced rapidly to address the interpretability limitations of DL models \cite{patricio2023explainable, guleria2026interpretable, zhang2026artificial}. Among the most widely used methods are Class Activation Mapping (CAM) and its gradient-based variant Grad-CAM, which produce heatmaps that highlight image regions influencing model predictions. Other gradient-based methods include Integrated Gradients, which attribute predictions to input features by accumulating gradients along a path from a baseline, and Guided Backpropagation, which refines gradient signals to generate sharper saliency maps. Perturbation-based methods such as SHAP and LIME estimate local model behaviour by modifying input features and measuring output changes to derive feature importance. Occlusion masks input regions to assess their impact, while sensitivity analysis quantifies output changes under input perturbations.

In breast cancer recognition, these XAI methods have been increasingly integrated with DL models to enhance clinical interpretability. Recent frameworks use Grad-CAM and variants such as Grad-CAM++ to visualise tumour regions in mammograms, ultrasound and histopathology images, supporting clinical validation of model attention against known lesions \cite{alom2025explainable, alhussen2025xai, zou2025explainable}. SHAP and LIME have also been applied to breast cancer classification models to provide feature-level explanations aligned with medical knowledge, enhancing trust and supporting clinical adoption \cite{peta2024explainable, murugan2025efficient}. Studies demonstrate that combining multiple XAI techniques can yield richer explanations. For instance, attention-guided Grad-CAM has been proposed to refine heatmaps by incorporating spatial and channel attention mechanisms for infrared breast thermography \cite{raghavan2024attention}. Quantitative evaluation commonly compares saliency maps with expert-annotated lesion masks using metrics such as localisation accuracy and overlap scores. Hybrid deep learning architectures that combine multiple CNN backbones with explainability modules improve both performance and interpretability. For example, fusion models integrating DenseNet121, Xception and VGG16 with GradCAM++ explanations achieve high accuracy while providing transparent visual justifications \cite{zou2025explainable}. Despite significant progress in this domain, there is a lack of studies that systematically evaluate existing XAI techniques.

\subsection{Deep Breast Cancer Recognition}
Recent years have seen substantial advances in DL for breast cancer recognition, with three primary architectural paradigms emerging as key directions: very deep neural networks, lightweight neural networks and transformer-based architectures \cite{shahid2025breast}. Each approach addresses distinct challenges in breast cancer recognition, with trade-offs in accuracy, computational efficiency and clinical applicability.

Very deep convolutional neural networks (CNNs), such as ResNet, DenseNet and VGG, have long served as the backbone of automated breast cancer detection systems. These architectures support hierarchical feature extraction from complex histopathological and radiological images and often achieve diagnostic performance comparable to expert radiologists \cite{singh2023deep, yuan2025comparative, nasir2025breast}. For instance, models such as ResNet50 and DenseNet121 achieve high accuracy in both binary and multi-class breast cancer classification across diverse datasets \cite{yuan2025comparative}. However, their depth and hierarchical representations limit traceability of how specific input features influence predictions \cite{allgaier2023does}. As a result, XAI outputs such as heatmaps or attributions tend to be less interpretable and less consistent with clinically meaningful patterns than those from simpler or more structured models.

To address these limitations, lightweight architectures, such as MobileNet, EfficientNet, SqueezeNet and compact CNN variants, have been developed. These models employ depthwise separable convolutions and bottleneck structures to reduce parameter counts while preserving competitive accuracy significantly. In particular, lightweight models are ideal for real-time clinical applications and resource-constrained environments. For instance, in \cite{kausar2023breast}, a lightweight CNN achieves strong classification performance in real-time breast cancer diagnosis. Similarly, several studies report comparable breast cancer detection performance with substantially reduced inference time and memory usage \cite{kubakisa2025automated, ma2025lmcnet, wang2025illuminating, mozebo2026enhancing}. However, their limited representational capacity can yield weaker or less detailed feature representations, which may reduce the clinical informativeness of explanations. In this context, XAI outputs may remain valid, but often appear less discriminative due to reduced modelling of complex patterns.

The emergence of transformer-based architectures has driven a paradigm shift in medical image analysis. Vision Transformers (ViT), Swin Transformers and hybrid CNN–transformer models employ self-attention to capture global context, addressing limitations of CNNs associated with local receptive fields. Very recently, transformer-based approaches have demonstrated strong performance in breast cancer recognition \cite{hayat2024hybrid, jahan2025deep, fatma2026magnification}. Attention maps can highlight regions that contribute to model decisions and offer a more intuitive interpretation than convolutional feature maps \cite{an2022attention, naas2025explainable}. However, the reliability of attention as an explanatory mechanism remains contested, as attention weights do not necessarily reflect true feature importance and can yield inconsistent interpretations.

Existing studies demonstrate substantial progress in DL-based breast cancer recognition and the integration of XAI techniques for improved explainability. However, current research mainly applies a limited set of explainability methods to individual DL architectures, with evaluations often restricted to qualitative visual inspection or single-model analysis. A systematic comparison of benchmark XAI techniques across diverse DL architectures for breast cancer recognition remains underexplored. In particular, limited evidence exists on the behaviour of different XAI methods across very deep CNNs, lightweight CNNs, transformer-based networks and hybrid networks under a unified experimental setting. Furthermore, quantitative evaluation of XAI performance using multiple assessment metrics remains insufficiently investigated.

\section{Proposed Method}
\label{sec:proposedmethod}
In this section, (i) the problem statement, (ii) representative and widely used XAI techniques and benchmark DL models employed and (iii) the dataset and evaluation framework are discussed.

\subsection{Problem Formulation}
This study uses breast ultrasound data. Let the breast ultrasound dataset be defined as:
\begin{equation}
\mathcal{D}_{breast} = \{(x_i, y_i, m_i)\}_{i=1}^{N}
\end{equation}
where $x_i \in \mathbb{R}^{H \times W \times C}$ denotes an input image, $y_i \in \{0,1\}$ represents the corresponding diagnostic label, and $m_i \in \{0,1\}^{H \times W}$ is the ground-truth lesion mask indicating clinically relevant regions annotated by experts.

In this study, four categories of DL architectures are considered, namely very deep neural networks, lightweight neural networks and transformer-based models. Let this set of architectures be represented as:
\begin{equation}
\mathcal{F} = \{f^{(d)}_{\theta}, f^{(l)}_{\theta}, f^{(t)}_{\theta}, f^{(h)}_{\theta}\}
\end{equation}
where $f^{(d)}_{\theta}$, $f^{(l)}_{\theta}$, and $f^{(t)}_{\theta}$ denote models from deep, lightweight, and transformer-based families, respectively, parameterised by $\theta$. Each model learns a mapping from the input image to a diagnostic prediction:
\begin{equation}
\hat{y}_i = f_{\theta}(x_i)
\end{equation}

To provide interpretability, a set of XAI techniques $\mathcal{E} = \{\mathcal{E}_1, \mathcal{E}_2, \dots, \mathcal{E}_K\}$ is applied to each trained model. For a given model $f_{\theta}$ and input $x_i$, an XAI method $\mathcal{E}_k$ generates a saliency defined as:
\begin{equation}
S_i^{(k)} = \mathcal{E}_k(f_{\theta}, x_i), \quad S_i^{(k)} \in \mathbb{R}^{H \times W}
\end{equation}
where each pixel value represents the importance of that region in contributing to the final prediction.

Since the objective is to evaluate the clinical relevance of explanations, the generated saliency maps are normalised to a common scale using min-max normalisation:
\begin{equation}
\tilde{S}_i^{(k)} = \frac{S_i^{(k)} - \min(S_i^{(k)})}{\max(S_i^{(k)}) - \min(S_i^{(k)})}
\end{equation}

To enable comparison with expert-annotated ground truth masks, a binarisation function $\tau(\cdot)$ is applied to obtain a predicted explanation mask:
\begin{equation}
\hat{m}_i^{(k)} = \tau(\tilde{S}_i^{(k)})
\end{equation}
where $\hat{m}_i^{(k)} \in \{0,1\}^{H \times W}$ represents the regions identified as important by the XAI method.

The central objective of this work is to evaluate how well the explanation masks $\hat{m}_i^{(k)}$ aligns with the clinically relevant regions $m_i$. This alignment reflects the faithfulness and clinical utility of the explanation method. Therefore, the overall explanatory performance of a method $E_k$ under a model $f_{\theta}$ is defined as:
\begin{equation}
\mathcal{R}(E_k, f_{\theta}) = \frac{1}{N} \sum_{i=1}^{N} \mathcal{M}\left(\hat{m}_i^{(k)}, m_i\right)
\end{equation}
where $\mathcal{M}(\cdot)$ denotes a set of evaluation metrics that quantify spatial agreement between the predicted explanation and ground-truth mask.

This study employs three complementary evaluation metrics to provide a comprehensive assessment of explanation quality from different perspectives. These metrics capture complementary aspects of explanation quality, including spatial agreement, localisation accuracy and region completeness, which provide a balanced and reliable evaluation. Intersection over Union (IoU) quantifies spatial agreement between the explanation and the ground truth by measuring their overlap relative to their union:
\begin{equation}
\mathrm{IoU} = \frac{|\hat{m}_i^{(k)} \cap m_i|}{|\hat{m}_i^{(k)} \cup m_i|}
\end{equation}
The Pointing Game (PG) evaluates localisation accuracy by verifying whether the most salient point of the explanation lies within the annotated region:
\begin{equation}
\mathrm{PG} = \mathbb{I}\left(\arg\max S_i^{(k)} \in m_i\right)
\end{equation}
Mean Coverage (MC) measures completeness by quantifying the proportion of the ground-truth region captured by the explanation:
\begin{equation}
\mathrm{MC} = \frac{|\hat{m}_i^{(k)} \cap m_i|}{|m_i|}
\end{equation}

\begin{figure}[!htbp]
    \centering
    \includegraphics[width=\linewidth]{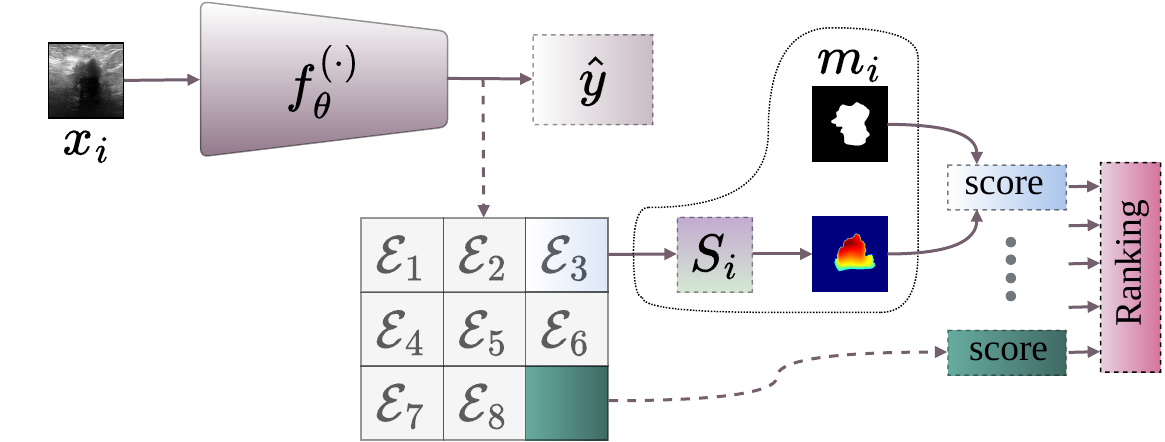}
    \caption{Illustration of the comparison of XAI output with the ground-truth lesion mask for a representative breast ultrasound image.}
    \label{fig:xaipipeline}
\end{figure}

Finally, the goal of this study is to determine the most suitable XAI method for each class of deep learning architecture. This can be formulated as an optimisation problem:
\begin{equation}
\mathcal{E}^{*}(f_{\theta}) = \arg\max_{\mathcal{E}_k \in \mathcal{E}} \mathcal{R}(\mathcal{E}_k, f_{\theta})
\end{equation}
which is subject to the constraint that $f_{\theta} \in \mathcal{F}$. This formulation enables a systematic and architecture-aware comparison of XAI techniques based on their ability to generate clinically meaningful explanations in breast cancer recognition, as shown in Figure \ref{fig:xaipipeline}. The figure presents a single example of saliency map generation and comparison with the corresponding mask. Similar analyses are conducted for each XAI technique to enable a comparative ranking of them.

\subsection{Benchmark XAI Techniques and DL Models}
To support a comprehensive and systematic evaluation of explainability methods in breast cancer diagnosis, this study incorporates nine widely used XAI techniques. These methods are selected to represent a diverse set of attribution paradigms, including activation-based, gradient-based, perturbation-based, surrogate-based and game-theoretic approaches, which ensures broad coverage of the main families of post-hoc explainability techniques. This selection enables a holistic assessment of interpretability across different explanatory paradigms for models $f_{\theta} \in \mathcal{F}$, evaluated on samples $x_i \in \mathcal{D}_{\text{breast}} = \{(x_i, y_i, m_i)\}_{i=1}^{N}$. The XAI techniques considered in this study are as follows:

\begin{enumerate}
    \item \textbf{CAM} \cite{zhou2016learning} produces class-discriminative localisation maps by leveraging weighted combinations of convolutional feature activations from the final convolutional layer. The resulting explanations highlight coarse but spatially meaningful regions that are most influential for the predicted class.

    \item \textbf{Grad-CAM} \cite{selvaraju2017grad} extends CAM by incorporating gradient information propagated from the predicted class score with respect to intermediate feature maps. This allows Grad-CAM to produce more task-sensitive localisation maps that better reflect the contribution of spatial regions to the decision.

    \item \textbf{Integrated Gradients (IG)} \cite{sundararajan2017axiomatic} attributes importance to individual input pixels in the images by accumulating gradients along a continuous path from a baseline input. This formulation ensures that attribution reflects the integrated sensitivity of the model output with respect to gradual changes in the input distribution.

    \item \textbf{SHAP} \cite{lundberg2017unified} provides feature attributions grounded in cooperative game theory by estimating the marginal contribution of each input feature to the prediction. SHAP ensures consistency and local accuracy, which makes it a theoretically grounded attribution method.

    \item \textbf{LIME} \cite{ribeiro2016should} constructs a locally faithful surrogate model by perturbing the input space and learning an interpretable approximation. This enables the explanation of individual predictions through a simplified linear or sparse representation.

    \item \textbf{Guided Backpropagation (GBP)} \cite{springenberg2014striving} modifies standard backpropagation by selectively propagating only positive gradients through activated neurons. This results in high-resolution saliency maps that emphasise fine-grained structures in the input, although without explicit class discrimination.

    \item \textbf{Saliency Maps (SM)} \cite{simonyan2013deep} compute first-order sensitivity of the model output with respect to the input image. This gradient-based approach provides a direct estimate of pixel-wise importance, which offers a simple yet effective visual explanation of model behaviour.

    \item \textbf{Occlusion} \cite{ancona2017towards} analysis evaluates explanation quality by systematically masking regions and measuring the resulting change in the model prediction. This perturbation-driven approach provides model-agnostic insights into spatial importance, albeit at a higher computational cost due to repeated inference.

    \item \textbf{Sensitivity} \cite{simonyan2013deep, ancona2017towards} analysis quantifies the robustness of explanations by measuring the stability of model gradients under small perturbations. This indicates how consistent attribution patterns remain under input noise.
    
\end{enumerate}
To ensure a comprehensive evaluation across four primary architectural paradigms, namely very deep neural networks, lightweight neural networks, transformer-based models and hybrid neural networks, a diverse set of models is selected. 

Very deep neural network models are characterised by high parameter capacity and deep hierarchical feature extraction, which support rich representation learning for complex visual patterns in breast ultrasound images.

\begin{enumerate}
    \item \textbf{ResNet50} utilises residual connections to enable stable training of deep architectures and effective feature propagation. This has been widely used in breast cancer recognition \cite{alzoubi2026enhancing}.

    \item \textbf{DenseNet} employs dense connectivity between layers to encourage feature reuse and strengthen gradient flow, which facilitates the extraction of discriminative features useful for breast cancer recognition \cite{sun2026enhanced}.
\end{enumerate}
Lightweight neural network models are optimised for computational efficiency, with reduced parameter counts and lower floating-point operations (FLOPs), which makes them suitable for real-time and edge-based clinical applications.

\begin{enumerate}
    \setcounter{enumi}{2} 
    \item \textbf{MobileNetV2} introduces inverted residual blocks and depthwise separable convolutions for efficient feature extraction. This lightweight design supports robust representation learning with reduced computational cost and is suitable for breast cancer recognition in resource-constrained clinical settings \cite{inamdar2026automated}.
    
    \item \textbf{ShuffleNet} utilises channel shuffling and grouped convolutions to reduce computational cost while maintaining accuracy. This efficient architecture supports effective feature learning with low latency and is suitable for breast cancer recognition in real-time clinical settings \cite{khati2026reciprocal}.
    
\end{enumerate}
Transformer-based neural networks use self-attention to model global dependencies and contextual relationships across the image, which supports the capture of complex structures in breast cancer recognition.

\begin{enumerate}
    \setcounter{enumi}{4}
    \item \textbf{Vision Transformer (ViT):} processes images as sequences of patches and applies global self-attention for representation learning. This global modelling capability is effective for capturing subtle and spatially distributed tumour patterns in breast ultrasound images, which are often difficult to localise using local receptive fields alone  \cite{naas2025explainable}.
    
    \item \textbf{Swin Transformer:} introduces hierarchical feature learning with shifted window-based self-attention for improved efficiency and scalability. This hierarchical representation is effective for breast cancer recognition, capturing both fine-grained lesion details and broader contextual tissue structures across multiple spatial resolutions \cite{aldawsari2026deep}.

\end{enumerate}
Hybrid neural networks integrate convolutional feature extraction with transformer-based self-attention, which enables the modelling of local spatial patterns and global contextual dependencies across the image. This supports the capture of complex multi-scale structures in breast cancer recognition.

\begin{enumerate}
    \setcounter{enumi}{6}
    \item \textbf{MobileViT:} A hybrid architecture that integrates lightweight convolutional operations with transformer blocks to capture both local and global representations, while maintaining efficiency constraints. This combination is effective for breast cancer recognition, supporting simultaneous modelling of fine-grained lesion boundaries and broader contextual tissue variations in ultrasound images \cite{potsangbam2025emvit}.
        
    \item \textbf{ConvNeXt} is relatively a new convolutional architecture that incorporates design principles inspired by transformer models while retaining convolutional efficiency. This design supports modelling of both local and global features, which is beneficial for capturing subtle structural patterns in breast cancer recognition \cite{kaur2026convnext}.

    \item \textbf{CoAtNet} is a hybrid architecture that combines convolutional operations with attention mechanisms, designed for high-capacity representation learning. This integration is beneficial for breast cancer recognition, improving joint modelling of fine-grained tumour textures and long-range contextual dependencies in heterogeneous breast tissue structures \cite{nivetha2025automated}.
\end{enumerate}

This categorisation supports an architecture-aware evaluation of XAI techniques across models with varying capacity, complexity and inductive biases, revealing links between model design and explainability in breast cancer diagnosis.

\subsection{Dataset and Evaluation Framework}
\begin{figure}[!htbp]
    \centering
    \includegraphics[width=\linewidth]{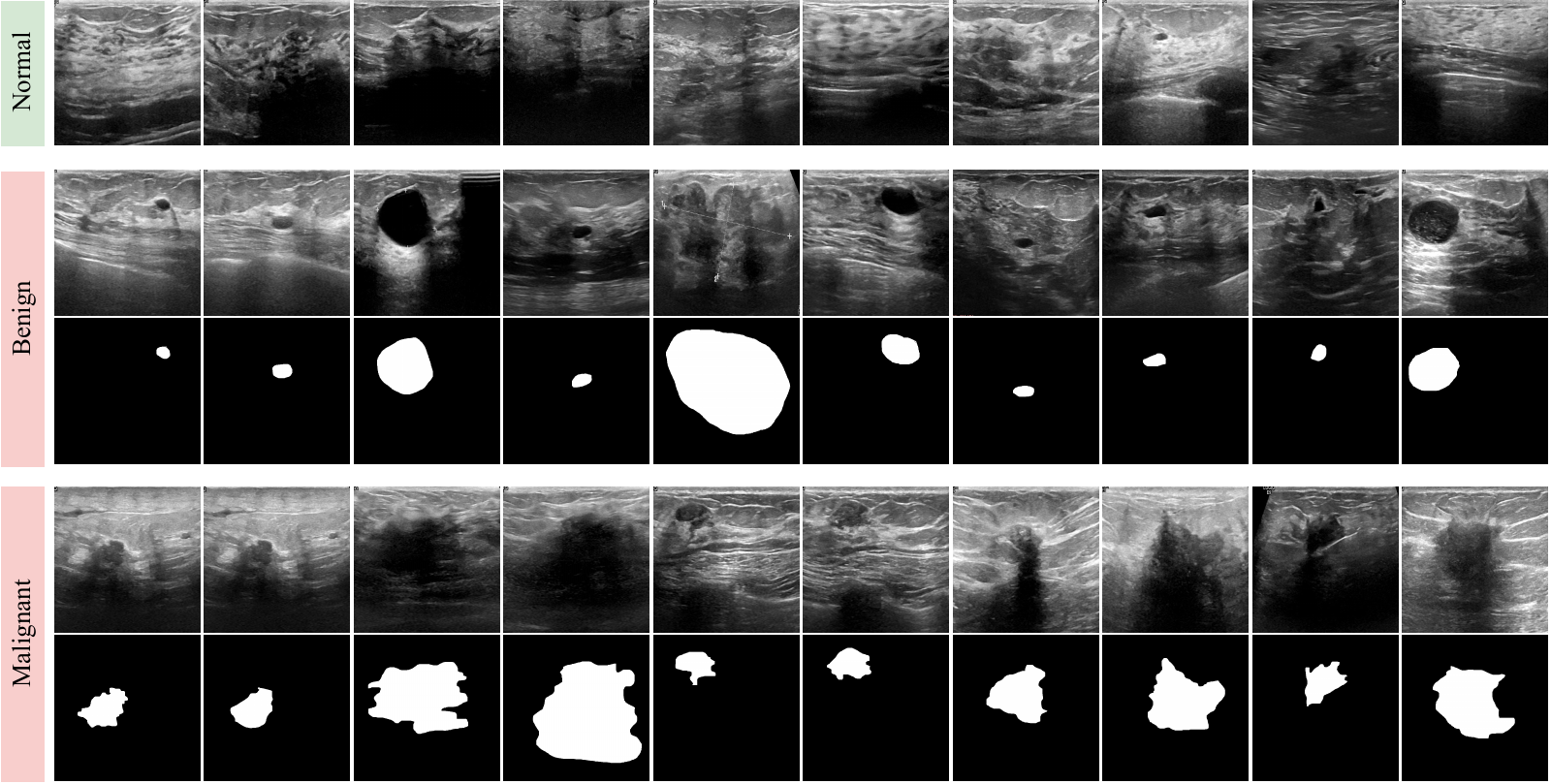}
    \caption{Representative samples from each class in the dataset, with corresponding ground-truth masks for benign and malignant cases shown in the second row.}
    \label{fig:dataset}
\end{figure}

This study uses the \textbf{Breast Ultrasound Dataset} \cite{al2020dataset}, a widely used publicly available dataset consisting of medical images developed to support machine learning tasks in breast cancer classification, detection and segmentation. The data was collected in 2018 at Baheya Hospital for early detection and treatment of women's cancer in Cairo, Egypt. The dataset is categorised into three distinct classes representing different clinical states, such as \textit{Normal}, \textit{Benign} and \textit{Malignant}. Figure \ref{fig:dataset} presents representative samples from each class. For the benign and malignant classes, the corresponding masks are shown in the second row. The dataset comprises 780 images across these three classes: 487 benign samples, 210 malignant samples and 133 normal samples. This study considers only the benign and malignant samples. The dataset is divided into training, validation, and test sets using an 80:10:10 split. To support segmentation-based analysis, each ultrasound image is paired with a corresponding ground-truth mask, which serves as the reference for evaluating the effectiveness of the XAI techniques in this study. These masks were created using freehand segmentation in MATLAB to define the precise boundaries of any detected masses.

Based on the problem formulation, an evaluation framework is designed to systematically assess the quality of explanations generated by different XAI techniques across multiple DL architectures for breast cancer recognition. The evaluation is conducted by forming all possible combinations between trained DL models and explainability methods. For each combination, explanations are generated for all test samples and assessed using standard metrics, such as IoU, PG and MC. To ensure robust assessment, metric scores are aggregated over the dataset by averaging across all samples. XAI methods are then ranked for each DL model based on their aggregated performance to identify the most suitable technique for each DL model. In addition, for each architecture category, results are aggregated across all models to enable a systematic comparison of explainability methods under different architectural paradigms.

\section{Experiments}
\label{sec:experiments}
This section presents the experimental setup, results for extensive experiments conducted to assess the proposed XAI evaluation framework and discussion and recommendations.

\subsection{Experimental Setup}
All experiments were implemented using the PyTorch\footnote{https://pytorch.org/} framework on a high-performance server equipped with two NVIDIA RTX A6000 GPUs, each with 49 GB of memory. The computing environment used NVIDIA driver version 550.144.03 with CUDA 12.4 support. The DL models used in this study were pre-trained on ImageNet \cite{deng2009imagenet} before being applied to breast cancer recognition and subsequently evaluated for explainability using XAI techniques. Training was performed using the AdamW optimiser with a learning rate of 0.0001 and weight decay values between 0.0003 and 0.0005, with a batch size of 16. Most models were trained for up to 300 epochs, while CoAtNet-0 and MobileViT-S were trained for 100 and 35 epochs, respectively. Regularisation techniques, including dropout (ranging from 0.2 to 0.4) and label smoothing (0.02–0.05), were consistently incorporated to enhance generalisation. A cosine annealing learning rate scheduler was additionally used for CoAtNet-0. To mitigate overfitting, early stopping with varying patience levels (8–20 epochs) was applied across all models, and the best-performing checkpoints were selected based on validation metrics. All these are selected empirically.

\subsection{Results}
The experimental results are presented in two stages. First, a performance evaluation of the DL models for breast cancer recognition. Second, a comparative analysis of XAI techniques to assess their effectiveness in model interpretation.

\subsubsection{DL Model Performance}
Table \ref{tab:dlcomparison} reports accuracy (A), precision (P), recall (R), and F1-score (F1) for the selected DL models, providing a structured comparison across architectural categories. The models are grouped into very deep, lightweight, transformer-based, and hybrid architectures to provide a clear comparison of their performance characteristics. 

The very deep neural networks, such as ResNet50 and DenseNet, demonstrate strong performance, with ResNet50 achieving an accuracy of 0.9367 and an F1-score of 0.9371, indicating effective capture of complex feature representations. Lightweight neural networks, such as MobileNetV2 and ShuffleNet, show slightly lower yet competitive performance, reflecting an efficiency–performance trade-off. Transformer-based neural networks such as ViT and Swin Transformer achieve similar results, indicating stable performance without a clear advantage in this setting. Hybrid neural networks show the most promising performance. ConvNeXt achieves the highest accuracy of 0.9494 with an F1-score of 0.9494, outperforming other neural networks, while CoAtNet also delivers strong results. Although MobileViT records the lowest performance in all neural networks, the overall results indicate that hybrid architectures provide a favourable balance between representation capacity and generalisation. These findings indicate that combining convolutional and attention mechanisms yields more robust performance than either approach alone.

\begin{table}
\centering
\caption{Comparison of deep learning models in terms of accuracy (A), precision (P), recall (R) and F1-score (F1). The numbers highlighted in blue and red indicate the best and least performing neural networks, respectively.}
\label{tab:dlcomparison}
\renewcommand{\arraystretch}{1.3}
\setlength{\tabcolsep}{6pt}

\begin{tabular}{|l|
                c|c|c|c|}
\hline
\textbf{Model} & 
\textbf{A} & 
\textbf{P} & 
\textbf{R} & 
\textbf{F1} \\
\hline

\multicolumn{5}{|c|}{\textbf{Very Deep Neural Networks}} \\ \hline
ResNet50         & 0.9367 & 0.9386 & 0.9367 & 0.9371 \\
DenseNet         & 0.9114 & 0.9130 & 0.9114 & 0.9117 \\

\hline
\multicolumn{5}{|c|}{\textbf{Lightweight Neural Networks}} \\ \hline
MobileNetV2      & 0.9114 & 0.9115 & 0.9114 & 0.9113 \\
ShuffleNet       & 0.9114 & 0.9143 & 0.9114 & 0.9122 \\

\hline
\multicolumn{5}{|c|}{\textbf{Transformer-based Neural Networks}} \\ \hline
ViT              & 0.9114 & 0.9189 & 0.9114 & 0.9129 \\
Swin Transformer & 0.9114 & 0.9130 & 0.9114 & 0.9121 \\

\hline
\multicolumn{5}{|c|}{\textbf{Hybrid Neural Networks}} \\ \hline
MobileViT        & \textcolor{red}{0.8987} & \textcolor{red}{0.9018} & \textcolor{red}{0.8987} & \textcolor{red}{0.9000} \\
ConvNeXt         & \textcolor{blue}{0.9494} & \textcolor{blue}{0.9510} & \textcolor{blue}{0.9494} & \textcolor{blue}{0.9494} \\
CoAtNet          & 0.9241 & 0.9269 & 0.9241 & 0.9256 \\

\hline
\end{tabular}
\end{table}

\begin{figure}[!htbp]
    \centering
    \includegraphics[width=\linewidth]{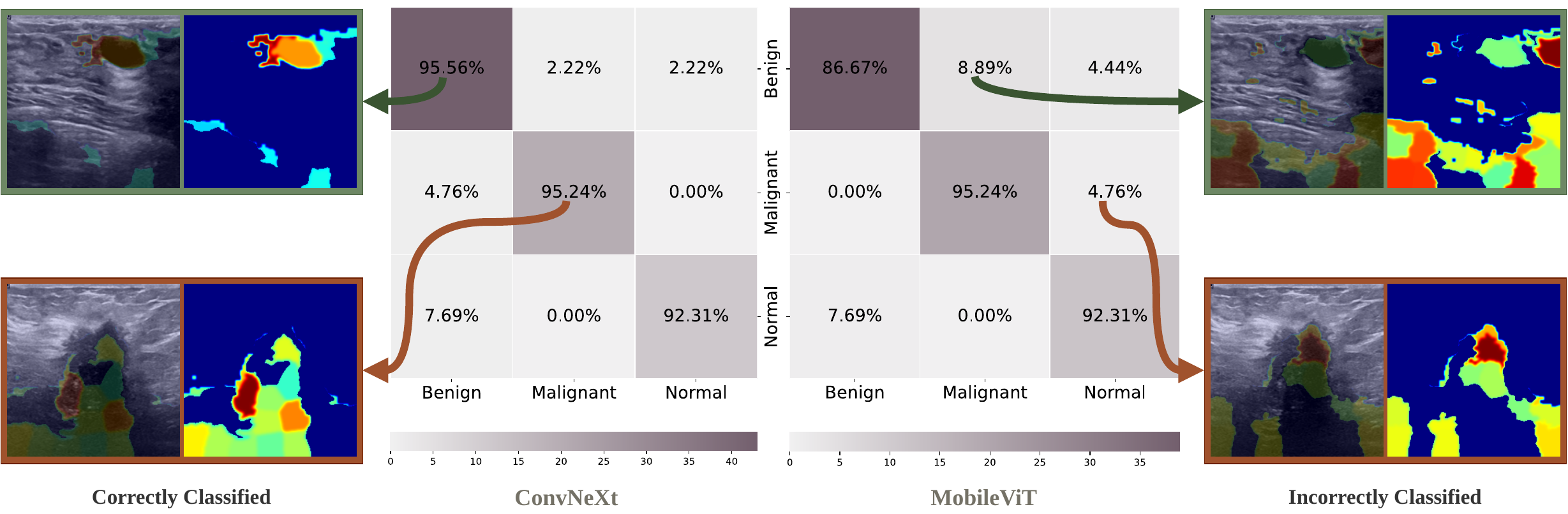}
    \caption{Qualitative comparison of XAI-based explanations for correctly and incorrectly classified breast cancer samples using ConvNeXt and MobileViT.}
    \label{fig:correctincorrect}
\end{figure}

\begin{table*}[htbp]
\centering
\caption{Detailed evaluation results for XAI techniques across DL models using PG, IoU and MC. The best PG, IoU, and MC values for each XAI technique within each DL model are highlighted in blue, while the overall best values across all DL models are indicated in yellow.}
\label{tab:xaicomparison}
\renewcommand{\arraystretch}{1.2}
\setlength{\tabcolsep}{4pt}

\begin{tabular}{|p{2cm}|p{0.8cm}|*{9}{>{\centering\arraybackslash}p{1cm}|}p{1.5cm}|}
\hline
\textbf{Model} & \textbf{Metric} & \textbf{CAM} & \textbf{GCAM} & \textbf{IG} & \textbf{SHAP} & \textbf{LIME} & \textbf{GBP} & \textbf{SAL} & \textbf{OCC} & \textbf{SENS} & \textbf{Best XAI} \\
\hline

\multicolumn{12}{|c|}{\textbf{Very Deep Neural Networks}} \\ \hline
\multirow{3}{*}{ResNet50} 
 & PG   & 0.1324 & 0.2941 & 0.4118 & \textcolor{blue}{0.5294} & 0.0588 & 0.5147 & 0.1912 & 0.2353 & 0.2059 & SHAP \\
 & IoU & 0.0537 & 0.0766 & 0.1382 & \textcolor{blue}{0.2255} & 0.0666 & 0.0350 & 0.0586 & 0.0934 & 0.0400 & SHAP \\
 & MC   & 0.0997 & 0.1325 & 0.2155 & \textcolor{blue}{0.3412} & 0.2137 & 0.0371 & 0.0922 & 0.2426 & 0.0589 & SHAP \\
\hline

\multirow{3}{*}{DenseNet} 
 & PG   & 0.2206 & 0.1912 & \textcolor{blue}{0.4706} & \textcolor{blue}{0.4706} & 0.1176 & 0.2206 & 0.3235 & 0.1912 & 0.2206 & IG/SHAP \\
 & IoU & 0.1270 & 0.1245 & 0.0876 & \textcolor{blue}{0.1599} & 0.0655 & 0.0739 & 0.0991 & 0.1043 & 0.0467 & SHAP \\
 & MC   & 0.2735 & 0.2568 & 0.1478 & \colorbox{yellow}{\textcolor{blue}{0.3505}} & 0.2499 & 0.0846 & 0.1551 & 0.2339 & 0.0748 & SHAP \\
\hline

\multicolumn{12}{|c|}{\textbf{Lightweight Neural Networks}} \\ \hline
\multirow{3}{*}{MobileNetV2} 
 & PG   & 0.3824 & 0.3971 & 0.5000 & 0.5000 & 0.0735 & 0.4853 & \textcolor{blue}{0.5294} & 0.2794 & 0.3676 & SAL \\
 & IoU & 0.1436 & 0.1536 & 0.1897 & \textcolor{blue}{0.1955} & 0.0833 & 0.1391 & 0.1406 & 0.1404 & 0.0875 & SHAP \\
 & MC   & 0.2750 & 0.2987 & 0.2563 & \textcolor{blue}{0.3131} & 0.2427 & 0.1872 & 0.1907 & 0.2787 & 0.1289 & SHAP \\
\hline

\multirow{3}{*}{ShuffleNet} 
 & PG   & 0.2647 & 0.3382 & 0.5147 & 0.5441 & 0.1176 & \textcolor{blue}{0.5588} & 0.2794 & 0.0735 & 0.2206 & GBP \\
 & IoU & 0.1670 & 0.1527 & 0.1155 & \textcolor{blue}{0.2279} & 0.0581 & 0.0819 & 0.0706 & 0.0769 & 0.0343 & SHAP \\
 & MC   & 0.2872 & 0.2674 & 0.1544 & \textcolor{blue}{0.3265} & 0.1878 & 0.0899 & 0.1020 & 0.2143 & 0.0408 & SHAP \\
\hline

\multicolumn{12}{|c|}{\textbf{Transformer-based Neural Networks}} \\ \hline
\multirow{3}{*}{ViT} 
 & PG   & 0.2206 & 0.1324 & 0.4853 & \textcolor{blue}{0.5588} & 0.1029 & 0.3676 & 0.3824 & 0.2794 & 0.0735 & SHAP \\
 & IoU & 0.0900 & 0.0498 & 0.1429 & \textcolor{blue}{0.1574} & 0.0657 & 0.1069 & 0.1092 & 0.1301 & 0.0184 & SHAP \\
 & MC   & \textcolor{blue}{0.2243} & 0.0800 & 0.2038 & 0.2082 & 0.2229 & 0.1596 & 0.1649 & 0.2449 & 0.0275 & OCC \\
\hline

\multirow{3}{*}{Swin Transformer} 
 & PG   & 0.1176 & 0.1176 & 0.2500 & 0.3235 & 0.1029 & \textcolor{blue}{0.4118} & 0.3235 & 0.2206 & 0.2500 & GBP \\
 & IoU & 0.0549 & 0.0549 & 0.0887 & 0.1471 & 0.0673 & 0.1639 & \textcolor{blue}{0.1686} & 0.1272 & 0.0577 & SAL \\
 & MC   & 0.1535 & 0.1535 & 0.1526 & 0.2559 & \textcolor{blue}{0.2653} & 0.2552 & 0.2569 & 0.2389 & 0.0932 & LIME \\
\hline

\multicolumn{12}{|c|}{\textbf{Hybrid Neural Networks}} \\ \hline
\multirow{3}{*}{MobileViT} 
 & PG   & 0.1765 & 0.1765 & \textcolor{blue}{0.5735} & 0.4118 & 0.1324 & 0.5000 & 0.5588 & 0.2794 & 0.4559 & IG \\
 & IoU & 0.1113 & 0.1113 & 0.1540 & \textcolor{blue}{0.1857} & 0.0583 & 0.1635 & 0.1706 & 0.1072 & 0.1342 & SHAP \\
 & MC   & 0.2431 & 0.2431 & 0.2000 & 0.2373 & 0.1536 & 0.2338 & 0.2431 & \textcolor{blue}{0.2758} & 0.1823 & OCC \\
\hline

\multirow{3}{*}{ConvNeXt} 
 & PG   & 0.2059 & 0.1471 & \textcolor{blue}{0.4265} & 0.4118 & 0.1765 & 0.3824 & 0.4118 & 0.2500 & 0.2500 & IG \\
 & IoU & 0.1779 & 0.1195 & 0.1659 & \textcolor{blue}{0.1918} & 0.0779 & 0.1536 & 0.1558 & 0.1072 & 0.1169 & SHAP \\
 & MC   & 0.2657 & 0.1779 & 0.2501 & 0.2534 & 0.2379 & 0.2701 & \textcolor{blue}{0.2787} & 0.2440 & 0.2156 & SAL \\
\hline

\multirow{3}{*}{CoAtNet} 
 & PG   & 0.0735 & 0.3235 & 0.6618 & 0.5588 & 0.1324 & 0.6029 &  \colorbox{yellow}{\textcolor{blue}{0.6912}} & 0.4412 & 0.5000 & SAL \\
 & IoU & 0.0748 & 0.1724 & 0.2018 & \colorbox{yellow}{\textcolor{blue}{0.2386}} & 0.0631 & 0.1725 & 0.1797 & 0.1276 & 0.1156 & SHAP \\
 & MC   & 0.2031 & \textcolor{blue}{0.2984} & 0.2256 & 0.2726 & 0.1693 & 0.2134 & 0.2216 & 0.2349 & 0.1554 & GCAM \\
\hline
\multicolumn{12}{|l|}{
\footnotesize
\begin{tabular}[t]{@{}l@{}}
\textit{Abbreviations:} CAM: Class Activation Mapping, GCAM: Gradient-weighted CAM, IG: Integrated Gradients, SHAP: SHapley Additive \\ exPlanations, LIME: Local Interpretable Model-agnostic Explanation, GBP: Guided Backpropagation, SAL: Saliency, OCC: Occlusion \\ and SENS: Sensitivity.
\end{tabular}
} \\
\hline
\end{tabular}
\end{table*}

For any DL model, achieving good recognition performance is important, as it directly affects the reliability of XAI outcomes. To illustrate this relationship, as shown in Figure \ref{fig:correctincorrect}, two representative test samples are selected from the benign and malignant classes and evaluated using both a high-performing and a low-performing model. The ConvNeXt model, which achieves the best overall classification performance in this study, correctly identifies both samples. In this setting, the corresponding XAI visualisations provide class-discriminative explanations that are consistent with clinically relevant regions. In contrast, the MobileViT model, which exhibits the lowest classification performance, misclassifies the same benign and malignant samples as malignant and normal, respectively. Although XAI maps can still be generated, they are conditioned on incorrect model predictions. The highlighted regions, therefore, reflect the basis of a miscalibrated decision rather than pathology-specific evidence. This comparison shows that XAI explanations depend on model accuracy, with higher-performing models producing more clinically meaningful and trustworthy attributions, while weaker models may yield visually plausible but misleading patterns.

\subsubsection{Explainability Analysis}
For the explainability analysis, PG, IoU and MC scores are computed for nine benchmark XAI techniques across nine DL models. The results are reported in Table \ref{tab:xaicomparison}. The results show notable variability in the performance of XAI techniques across different DL architectures and evaluation metrics. 

Among very deep neural networks, SHAP consistently attains the highest PG, IoU and MC scores. This suggests that game-theoretic attribution methods produce more spatially aligned and comprehensive explanations for models with high representational capacity. In contrast, the gradient-based method IG achieves competitive PG scores but exhibits inconsistent IoU and MC performance, indicating limited ability to capture complete lesion regions. For lightweight neural networks, the performance landscape is more fragmented. SHAP again shows strong and consistent IoU and MC performance across MobileNetV2 and ShuffleNet, while other methods obtain the highest PG scores. This divergence indicates that explanations from lightweight models are sensitive to the choice of XAI technique, with no single method consistently dominating across all metrics. The lower and more variable scores reflect the reduced representational capacity of these models, which affects the quality and stability of the generated explanations.

Transformer-based models exhibit moderate and less consistent XAI performance. SHAP remains competitive in PG and IoU for ViT, while Occlusion and LIME achieve higher MC scores, indicating metric-dependent behaviour. Similarly, the Swin Transformer exhibits no clear dominant XAI method, with different techniques performing best across PG, IoU and MC. This inconsistency indicates that attention-based architectures do not inherently ensure superior explainability and their explanations remain dependent on the evaluation metric and the chosen method. Hybrid architectures exhibit the most consistent and robust performance across all categories. In particular, CoAtNet attains the highest overall PG and IoU scores, with Saliency and SHAP emerging as strong performers. ConvNeXt and MobileViT also demonstrate competitive performance, with Integrated Gradients and Saliency showing strong performance across multiple metrics. The strong performance of hybrid models suggests that combining convolutional and attention mechanisms yields more informative and spatially aligned explanations. 

\begin{figure}[!htbp]
    \centering
    \includegraphics[width=\linewidth]{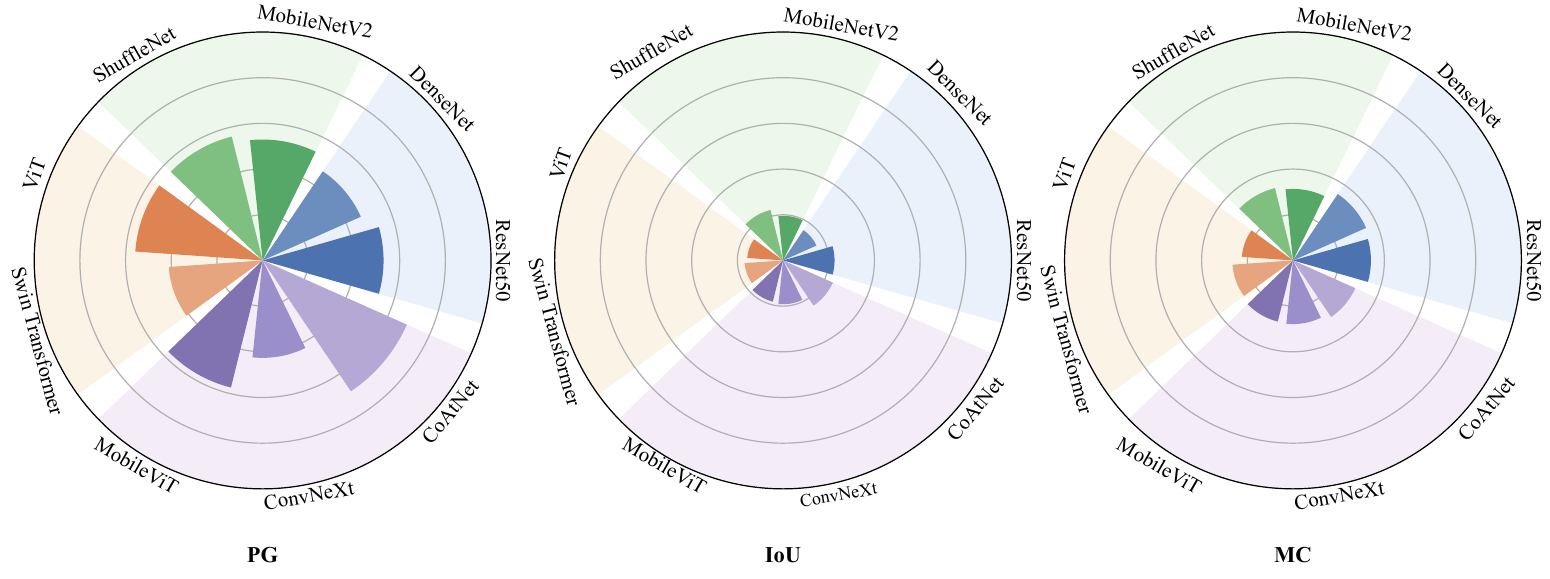}
    \caption{Best-performing XAI methods across models for PG, IoU and MC. The coloured backgrounds in the plot indicate the neural network categories.}
    \label{fig:bestxaicomparison}
\end{figure}

The results indicate that no single XAI technique consistently dominates, and that the effectiveness of explainability methods depends on the underlying model architecture and evaluation criterion. This trend is further illustrated in Figure \ref{fig:bestxaicomparison}, which presents a consolidated view of the best-performing XAI results across models for PG, IoU and MC. A notable pattern is the variation in metric sensitivity, with PG showing a wider spread than IoU and MC, indicating greater differentiation in localisation capability across models. In contrast, IoU and MC appear more compressed, suggesting that overlap- and coverage-based evaluations are less discriminative at higher performance levels. The plot highlights the importance of metric-specific behaviour in interpreting XAI performance, as different metrics capture distinct aspects of explanation quality.

\begin{figure}[!htbp]
    \centering
    \includegraphics[width=\linewidth]{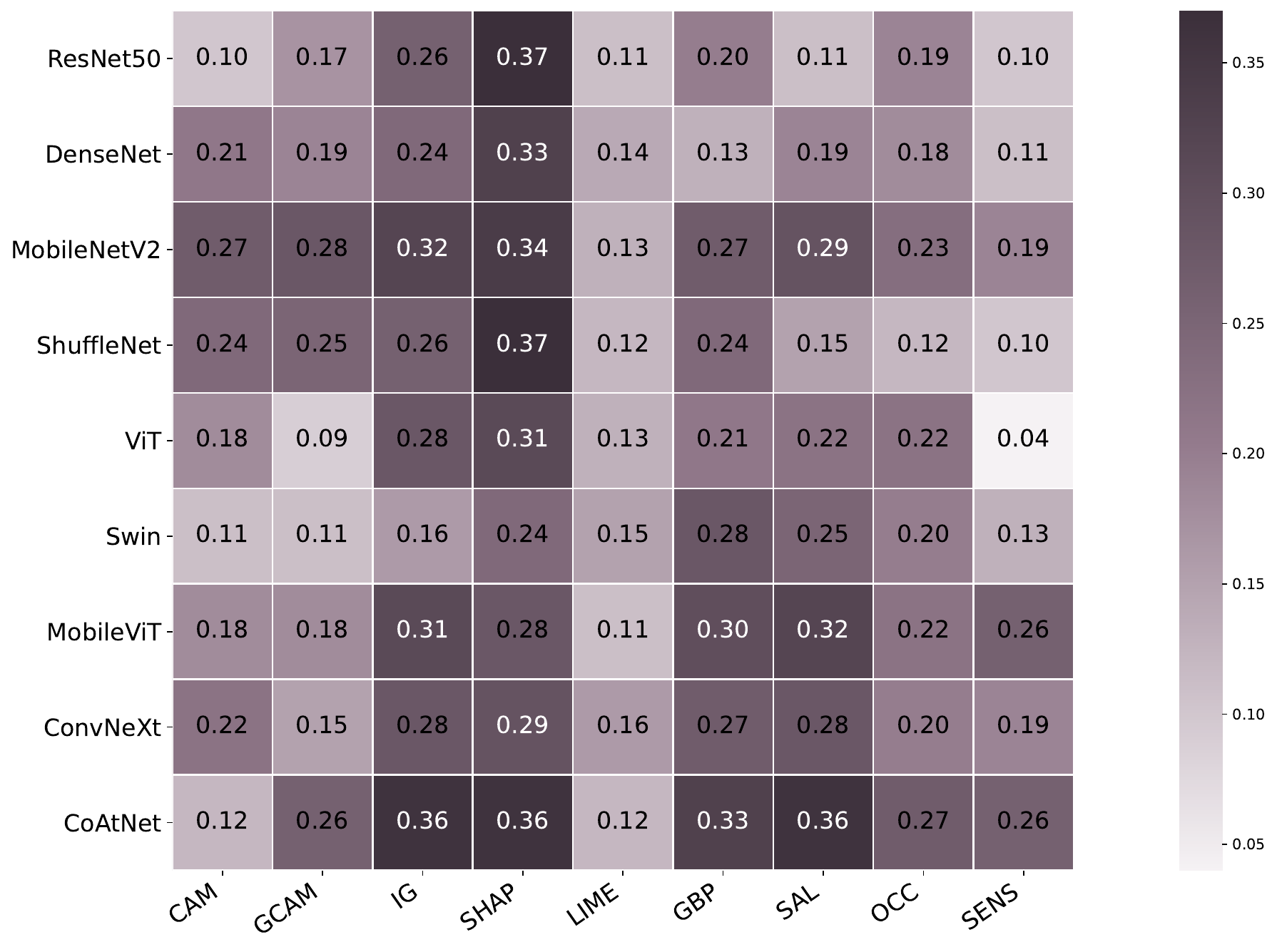}
    \caption{Heatmap of average PG, IoU and MC scores against individual DL models.}
    \label{fig:globalxaidistribution}
\end{figure}

Figure \ref{fig:globalxaidistribution} presents a heatmap of the average PG, IoU, and MC scores. It shows a highly non-uniform interaction-dependent distribution of XAI performance across nine deep learning models and nine explanation methods, indicating that explainability is driven by DL model–XAI technique compatibility rather than intrinsic method superiority. SHAP and IG consistently achieve stronger and more stable performance across most architectures, indicating greater robustness, whereas CAM-based and simpler perturbation methods, such as LIME and GBP, exhibit weaker and less consistent performance, particularly on transformer-based models. Hybrid neural networks achieve the most favourable attribution scores, while very deep neural networks remain stable but moderate, and transformer-based and lightweight models exhibit greater variability. Overall, the results indicate a strong coupling between architectural inductive bias and XAI effectiveness, supporting joint evaluation of DL models and XAI techniques rather than isolated assessment.

\begin{figure*}[!htbp]
    \centering
    \includegraphics[width=\linewidth]{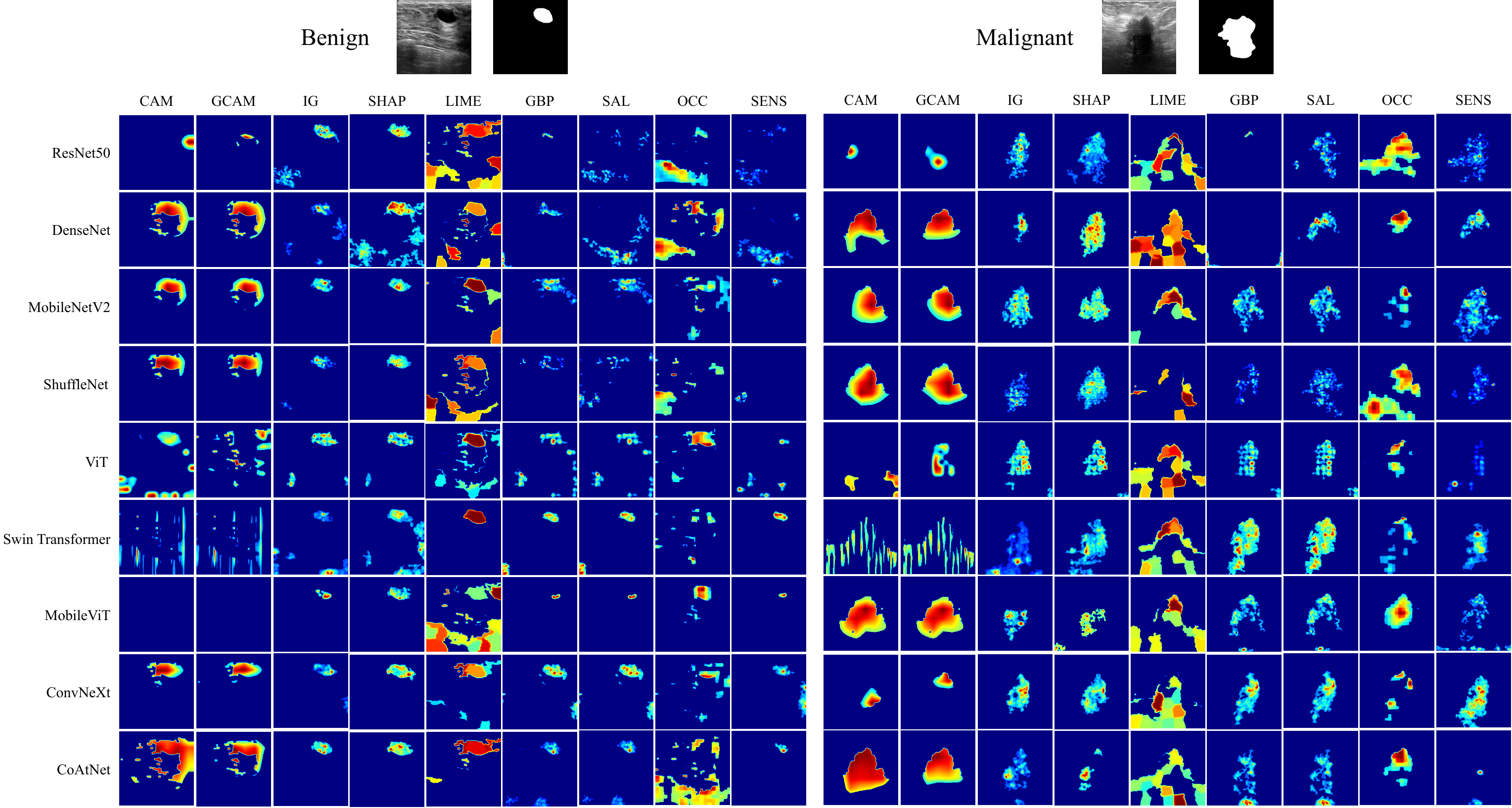}
    \caption{Qualitative illustration of the outputs generated by the XAI techniques for two representative samples taken from benign and malignant classes.}
    \label{fig:visualisexai}
\end{figure*}

Furthermore, in Figure \ref{fig:visualisexai}, a qualitative comparison of XAI-generated explanations across nine DL models for two representative samples, one benign and one malignant, is presented. For the benign case, high-performing neural networks such as CoAtNet produce explanations that are well localised within the annotated lesion boundaries, with saliency concentrated in clinically relevant regions. In contrast, lightweight neural networks and some transformer-based neral networks exhibit more fragmented attribution patterns that indicate weaker spatial alignment with the ground-truth mask. A similar trend is observed for the malignant sample, where hybrid models produce more coherent and concentrated activation regions that closely follow tumour structures. Neural networks with lower classification performance tend to highlight irrelevant regions or distribute attention across non-discriminative areas. These observations are consistent with the quantitative findings, which indicate that explanation quality is closely associated with both neural network architecture and predictive performance. The variability across neural networks indicates that different architectural categories capture distinct feature representations that influence the spatial characteristics of the generated explanations.

\subsection{Discussion and Recommendations}
The experimental findings indicate that explainability performance is primarily governed by the interaction between model architecture and attribution mechanism, rather than by the intrinsic superiority of any single XAI technique. This dependency indicates that the quality of the explanation is constrained by the feature representations learned by the underlying model. Architectures that capture both local texture and global contextual information tend to produce explanations with improved spatial coherence and clinical relevance. In contrast, models with less structured feature hierarchies produce explanations that are less stable and more sensitive to the choice of XAI method. This indicates that explainability should be treated as an emergent property of the joint model–method design, rather than as a post hoc component.

From a methodological perspective, the variation across evaluation metrics indicates that different XAI techniques optimise distinct aspects of explanation quality. Localisation-focused methods tend to prioritise peak attribution, whereas others favour broader region coverage or smoother spatial distributions. Reliance on a single evaluation metric may lead to biased assessments of explanation effectiveness. A multi-metric evaluation is necessary to capture complementary dimensions of explanation quality and to avoid overestimating the reliability of visually appealing but incomplete attributions. 

Based on these observations, several recommendations emerge for practical deployment. The selection of XAI techniques should be architecture-aware, prioritising methods that demonstrate stable performance across multiple evaluation criteria within a given model family. Hybrid architectures are better suited for clinical explainability applications, as they offer a balanced trade-off between predictive performance and interpretable feature representation. Explanation reliability should be evaluated alongside model correctness, since explanations from incorrect predictions may lead to misleading clinical interpretations. Future work should focus on integrated optimisation frameworks that jointly improve predictive accuracy and explanation faithfulness.

\subsection{Limitations}
Despite providing a systematic, architecture-aware evaluation of XAI techniques for breast cancer recognition, this study has two notable limitations.

The evaluation is conducted on a single breast ultrasound dataset of limited size from one clinical source. While suitable for controlled comparison, this limits generalisability across imaging modalities, acquisition protocols and patient populations. XAI behaviour is sensitive to data distribution and explanations that align well in one dataset may not remain consistent under domain shifts. Future work should include multi-centre, multi-modal datasets, such as mammography and histopathology, to assess robustness under varied clinical conditions. Cross-dataset validation and domain generalisation strategies can further evaluate the stability of architecture–XAI interactions across diverse settings.

The current evaluation relies on quantitative spatial metrics such as IoU, PG and MC. While these metrics assess spatial alignment, they do not capture clinical interpretability, diagnostic relevance, or trustworthiness. Explanations with high metric scores may still lack alignment with clinical reasoning. Incorporating expert-in-the-loop evaluation is necessary to address this gap. Future studies should involve radiologists in structured assessments of explanation usefulness, diagnostic consistency and trust. Combining quantitative metrics with qualitative clinical evaluation would provide a more comprehensive assessment of explainability.

\section{Conclusion}
\label{sec:conclusion}
This study presented a systematic and architecture-aware evaluation of XAI techniques for breast cancer recognition using DL models. A comprehensive experimental framework was developed to evaluate nine representative XAI methods across four architectural paradigms, such as very deep, lightweight, transformer-based and hybrid neural networks, using clinically aligned spatial evaluation metrics. The results indicate that explainability performance is not determined solely by the attribution method, but is strongly influenced by the underlying model architecture and its learned feature representations. In particular, hybrid architectures consistently produce more stable and clinically meaningful explanations because they capture both local and global contextual information, whereas other architectures exhibit greater variability in explanation quality depending on the XAI technique employed. Furthermore, the comparative analysis shows that no single XAI method consistently generalises across all neural networks and evaluation criteria, demonstrating the need to jointly consider model design and explanation strategy in medical imaging applications. Overall, the findings indicate that reliable interpretability results from the interaction between the neural network and the XAI method rather than functioning as an isolated component. This demonstrates the importance of selecting appropriate architecture–XAI combinations for trustworthy clinical decision support. Future work will extend this framework by incorporating systematic evaluation from medical experts to assess the clinical suitability and trustworthiness of XAI outputs, and by investigating multimodal explanation strategies that integrate visual attributions with complementary textual explanations to improve interpretability and practical usability in real-world diagnostic settings.

\section*{Acknowledgement}
All authors declare that they have no known conflicts of interest in terms of competing financial interests or personal relationships that could have an influence or are relevant to the work reported in this paper.

\bibliographystyle{IEEEtran}
\bibliography{refs}

\vfill

\end{document}